\documentclass[11pt]{article}

\usepackage{times}

\usepackage[final]{acl}

\usepackage{stfloats}
\usepackage[T1]{fontenc}    % use 8-bit T1 fonts
\usepackage{hyperref}       % hyperlinks
\usepackage{url}            % simple URL typesetting
\usepackage{booktabs}       % professional-quality tables
\usepackage{amsfonts}       % blackboard math symbols
\usepackage{nicefrac}       % compact symbols for 1/2, etc.
\usepackage{microtype}      % microtypography
\usepackage{xcolor}         % colors
\usepackage{array}
\usepackage{amssymb}
\usepackage{adjustbox}
\usepackage{multirow}
\usepackage{amssymb}
\usepackage{natbib}
\usepackage{tikz}
\usepackage{pgfplots}
\pgfplotsset{compat=1.18}

\title{Pula: Training Large Language Models for Setswana}

\author{
  Nathan Brown\\
  \textit{Data Science for Social Impact} \\
  University of Pretoria \\
  School of Computing\\
  Clemson University\\
  \textit{nbrown9@clemson.edu}\\
  \And
  Vukosi Marivate \\
  \textit{Data Science for Social Impact} \\
  Department of Computer Science \\
  University of Pretoria\\
  Lelapa AI\\
  \textit{vukosi.marivate@cs.up.ac.za}\\
}

\begin{document}

\maketitle

\begin{abstract}
In this work we present \textbf{Pula}, a suite of bilingual language models proficient in both Setswana and English. Leveraging recent advancements in data availability and efficient fine-tuning, Pula 8B and Pula 14B outperform GPT-4o and Gemini 1.5 Pro on English-Setswana translation tasks and achieve state-of-the-art performance on Setswana reasoning tasks for their size. We release the weights for Pula \textbf{1B}, \textbf{3B}, \textbf{8B}, and \textbf{14B} as well as training logs and training and evaluation code. Alongside Pula, we release the largest-ever Setswana text corpus, \textbf{Marothodi}, and the first comprehensive Setswana instruction-tuning dataset, \textbf{Medupi}, consisting of reformatted datasets, translated corpora, and synthetic LLM-generated text. To accompany this data, we release the code used for dataset construction, formatting, filtering, and scraping. Last, we release two Setswana LLM-translated benchmarks, \textbf{MMLU-tsn} and \textbf{GSM8K-tsn}, to measure Setswana knowledge and reasoning capabilities.
\end{abstract}

\section{Introduction}

Setswana, also known as Tswana, is a Bantu language spoken by an estimated five to ten million people worldwide \cite{Bennett_Diemer_Kerford_Probert_Wesi_2016}. Closely related to Northern Sotho and Southern Sotho, Setswana holds official status in Botswana \cite{botswana_gov}, South Africa \cite{southafrica_constitution}, and Zimbabwe \cite{zimbabwe_constitution}, and is also used in countries like Namibia, often interchangeably with English \cite{norris2017sound}. Despite its significance in the lives of millions, Setswana has been largely overlooked in natural language processing (NLP) research, despite being classified by some works as a high-resource language \cite{nllb}. This work aims to bridge the gap between Setswana and other high-resource languages by making open generative large language models capable of high-quality Setswana available to the community for the first time, significantly increasing data availability, and laying the groundwork for future Setswana-centric research.

Large Language Models (LLMs) have demonstrated powerful capabilities across various domains after training on web and synthetic data \cite{gpt4, llama, claudesonnet, textbooksareallyouneed}, excelling in areas such as mathematics \cite{mathΣtral}, programming \cite{deepseekcoder}, creative writing \cite{weaver}, and translation tasks \cite{attentionisallyouneed}. However, developers continue to primarily target English and certain high-resource languages in training and evaluation. While existing approaches yield impressive capabilities, they may produce models which lack knowledge of certain cultures, limit production use-cases outside majority demographics, or prevent a significant portion of the global population from utilizing language models effectively. African languages like Setswana, with little textual data available compared to languages such as English or French, subsequently suffer in performance and are underutilized in research.

Recent progress has been made to address the lack of language diversity in language models. Releases such as mBART \cite{mbart}, XLM-RoBERTa \cite{xlmr}, and BLOOM \cite{bloom} were among the earliest and most influential advancements in multilingual language models. Building upon these technologies, newer models including GPT-4 \cite{gpt4}, Claude \cite{claudesonnet}, Gemini \cite{gemini}, Llama \cite{llama}, and Gemma \cite{gemma} have also found success in multilingual domains, often demonstrating reasoning and translation capabilities in languages not officially supported. Releases such as Aya 101 \cite{aya101} and Aya 23 \cite{aya23} have continued to improve language coverage for translation and generative tasks, and open corpora such as ROOTS \cite{roots}, OSCAR \cite{oscar}, and mC4 \cite{mc4} have made multilingual pre-training data readily available. However, Setswana comprises only a small fraction of these datasets. For instance, just 0.0002\% of the ROOTS corpus is written in Setswana. Moreover, much of the available Setswana text in open multilingual corpora is of lower quality or predominantly religious in nature, resulting in significantly worse conversational, translation, and reasoning capabilities in current open models.

To help address this issue, we introduce the \textbf{Pula} series of language models. This series consists of LoRA \cite{lora} and QLoRA \cite{qlora} fine-tuned versions of Llama 3.2 1B, Llama 3.2 3B, Llama 3.1 8B, and Qwen 2.5 14B \cite{qwen2, llama}. By training a range of models across different parameter counts, we aim to provide the research community and millions of Setswana speakers with models that are both highly performant and capable of running on various hardware configurations ranging from data centers to consumer laptops and mobile phones. To fuel the training behind these models and to provide the research community with resources to further improve future models and research, we hand-curate the largest-ever corpus of Setswana text. In doing so, we merge several existing corpora which have not yet been consolidated, restore document-level text in certain data subsets, locate new sources of Setswana text which to our knowledge have not yet been utilized in NLP research, develop several dedicated scrapers and parsers to obtain this new data, reformat existing datasets to function as instruction-tuning corpora, translate existing English instruction-tuning datasets to Setswana using state-of-the-art large language models and translation models, and utilize GPT-4o for the generation of entirely new synthetic Setswana text. We name our pre-training dataset \textbf{Marothodi} and our instruction-tuning dataset \textbf{Medupi}.

Following the open approach of projects like OLMo \cite{olmo} and Dolma \cite{dolma}, we perform a fully open release. This includes model weights, all training data, metadata on data sources, training logs, and code for training, evaluation, dataset curation, dataset translation, synthetic data generation, and web scraping. All model weights and data can be accessed on \href{https://huggingface.co/collections/OxxoCodes/pula-66af1106ccc0fb38839f39da}{Hugging Face}~\footnote{\url{https://huggingface.co/collections/OxxoCodes/pula-66af1106ccc0fb38839f39da}}. All code and training logs can be accessed on \href{https://github.com/OxxoCodes/Pula}{GitHub}~\footnote{\url{https://github.com/OxxoCodes/Pula}}.

\section{Related Work}

Although much of current NLP research is English-centric, there have been significant recent advancements in Setswana-centric language models and data access. TswanaBERT \cite{tswanabert} represents one of the earliest examples, trained on over ten thousand Setswana sentences from the Leipzig Corpora Collection \cite{leipzig}, SABC news headlines \cite{sabcnewsheadlines}, and various blogs and websites. More recently, the NCHLT Setswana RoBERTa model \cite{nchltroberta} was released, having been trained on over fourteen million tokens of Setswana text from the NCHLT \cite{nchlt}, Autshumato \cite{autshumato}, Leipzig \cite{leipzig}, and Common Crawl corpora, and an internal CTexT corpus. PuoBERTa \cite{puoberta} marked a significant step forward in masked language modeling, achieving state-of-the-art performance while being the first language model trained from scratch. PuoBERTa was released alongside PuoData, the largest collection of curated Setswana text at the time, totaling 4.5 million PuoBERTa tokens excluding JW300 \cite{jw300}.

Much of the literature on African languages has targeted massively multilingual developments with a focus on machine translation and transfer learning. Corpora such as OPUS \cite{opus} and MADLAD-400 \cite{madlad400} provide access to large volumes of parallel web text data, while work such as MAFAND-MT \cite{mafand-mt} has enabled improved translation performance across many African languages through additional human-generated data. Meta's No Language Left Behind (NLLB) \cite{nllb} has facilitated high-quality machine translation between over 200 languages, including Setswana, although it is noted as one of the 21 languages with the lowest accuracy on FLORES-200 \cite{flores101}. MADLAD-400 \cite{madlad400} has allowed for increased multilinguality, but we find suffers from worse performance on Setswana-English translations as seen in Table \ref{tab:performance}. Furthermore, models such as AfriBERTa \cite{afriberta} and AfroXLMR \cite{afroxlmr} have seen success in training masked language models across multiple African and low-resource languages, and LlamaX \cite{llamax} has demonstrated high degrees of translation performance across over 100 languages as a LLM while retaining generalization.

Thanks to the cumulative improvements made by the research community in Setswana-centric NLP and increased levels of multilinguality, we believe there are significant opportunities to enhance existing Setswana NLP systems. Through our investigations we find several smaller corpora of Setswana text are often underutilized in the literature, potentially due to differing distribution methods and vastly differing formats, making data difficult to locate and utilize. For example, the South African Center for Digital Language Resources~\footnote{\url{https://sadilar.org}} \cite{sadilar} has made publicly available several datasets of Setswana text and audio. However, these texts are not distributed in a standardized format and are not typically available on commonly used external platforms such as Hugging Face and Kaggle. As such, much of this data is left out of existing corpora. We also find many websites hosting content written in Setswana are excluded from existing training datasets, meaning a significant portion of the available high-quality, and especially educational, Setswana data is not being utilized. Through the development and release of Marothodi and Medupi, we aim to reduce this burden and make data significantly easier to access.

\definecolor{myorange}{HTML}{FE9A09}
\definecolor{white}{HTML}{FFFFFF}
\begin{figure}[h]
    \centering
    \begin{tikzpicture}
    \begin{axis}[
        width=\linewidth,
        ybar,
        ymode=log,          % Set the y-axis to logarithmic scale
        log basis y=10,     % Optional: sets the logarithm base to 10
        ylabel={Total Tokens (Millions)},
        symbolic x coords={\textbf{Medupi}, OPUS, \textbf{Marothodi}, MADLAD-400, PuoData, Leipzig, GlotCC V1.0},
        xtick=data,
        x tick label style={rotate=45,anchor=east},
        ymin=0.5,           % Use a small positive value instead of 0
        scale=1.0
    ]
        \addplot[fill=myorange, draw=myorange] coordinates {
            (\textbf{Medupi}, 246.500255)
            (OPUS, 210.548909)
            (\textbf{Marothodi}, 73.952642)
            (MADLAD-400, 10.952546)
            (PuoData, 8.537662)
            (Leipzig, 1.931069)
            (GlotCC V1.0, 0.941585)
        };
    \end{axis}
    \end{tikzpicture}
    \caption{\centering Logarithmic comparison of total tokens per corpus, as measured using Llama 3.1's tokenizer.}
    \label{tab:token_counts}
\end{figure}

\section{Data}

One key consideration in Pula's design and subsequent training data selection is identifying target languages. There are several parallel corpora available for Setswana, and some works such as OPUS contain Setswana text paired alongside multiple other languages. However, significantly more English-Setswana parallel text is available. For example, OPUS contains over four times more parallel Setswana-English sentences than Setswana-French sentences~\footnote{\url{https://opus.nlpl.eu/results/tn&en/corpus-result-table}}. In addition, many publicly available, commonly used, high-quality, educational, and instruction-tuning datasets are written predominantly in English. Consequently, many state-of-the-art LLMs already excel in English tasks, including the Llama and Qwen series of models which we select for continued pre-training. Additionally, native speakers in regions such as Botswana and South Africa often utilize English in legal, official, and government documents, with English sometimes being spoken interchangeably with Setswana \cite{hansard}. We observe this trend continued in web text documents, where many data sources are written either exclusively in Setswana or in both Setswana and English, either incorporating code-switching or providing direct translations \cite{setswanamobotswana}.

Due to the strong link between these two languages and in an effort to reduce scope and computational requirements, we focus our efforts primarily on curating a high-quality dataset consisting of Setswana and English texts. We refer to our corpus of raw Setswana text as \textbf{Marothodi}, and our instruction-tuning dataset as \textbf{Medupi}.

Our datasets build upon several prior works in the African NLP research community. We measure token counts using the Llama 3.1 tokenizer throughout this paper for consistency. We release both Marothodi and Medupi in their entirety on Hugging Face~\footnote{\url{https://huggingface.co/collections/OxxoCodes/pula-66af1106ccc0fb38839f39da}}, as well as per-sequence identification metadata including a corpus identifier and an exact source URL where available.

\begin{table*}[h!]
\centering
\begin{adjustbox}{max width=\linewidth}
\begin{tabular}{@{}llll@{}}
\toprule
\textbf{Source} & \textbf{Data Category} & \textbf{Tokens} & \textbf{Description} \\
\midrule
Other Corpora & Existing Datasets & 21,783,242 & Bloom-lm, GlotCC, MADLAD-400, Vuk'uzenzele, etc. \\
Educational Material & Web Scraping & 21,567,283 & Exams, Quizzes, Books \\
TinyStories-tsn & Translated & 13,327,017 & Setswana translated subset of TinyStories \\
SADiLaR & Existing Datasets & 4,770,707 & Transcripts, ASR/TTS \\
Setswana Rare Books & Web Scraping/OCR & 4,428,724 & Old books, high quality \\
Governmental Documents & Web Scraping & 4,222,174 & Legal texts \\
Setswana Bible & Web Scraping & 1,712,827 & Religious texts \\
Setswana Wikipedia & Restored & 1,043,992 & Broad knowledge, high quality \\
Miscellaneous Websites & Web Scraping & 565,562 & iAfrika, Setswana Mo Botswana, Tlhalefang, Unisa \\
Miscellaneous Documents & Web Scraping & 327,429 & Declaration of Human Rights, Intro to Setswana, etc. \\
Nalibali & Restored, PDF Extraction & 203,685 & Children's stories, educational \\
\bottomrule
\end{tabular}
\end{adjustbox}
\caption{\centering{Key Data Sources in Marothodi.}}
\label{tab:marothodi_sources}
\end{table*}

\subsection{Marothodi}

\paragraph{Motivation.}
Marothodi is built with two goals in mind:

\begin{enumerate}
    \item To dramatically increase the amount of publicly available Setswana text through manual curation and filtering.
    \item To significantly increase ease of access for the African NLP research community and the general public.
\end{enumerate}

We believe a significant limitation in Setswana-centric NLP is the sparsity of published data: resources are scattered across various publishers such as Hugging Face, SADiLaR, various massive web crawls, public websites, and cloud storage providers. As a result, researchers who are not deeply familiar with the current Setswana data landscape or do not have ample time to allocate toward data collection are severely limited in data diversity, quality, and quantity. To assist in resolving this problem, we develop Marothodi: a Setswana pre-training dataset pulling together data from 33 different sources totaling 74 million tokens. Marothodi contains nearly 9 times as many tokens as are present in PuoData, making Marothodi the largest available single corpus of Setswana text.

\paragraph{Data Sources.}
We develop Marothodi by selectively including various underutilized data sources. We prioritize document integrity and rich context by utilizing complete documents from various sources. To aid in this effort, we restore the original documents of certain subsets of PuoData, including Wikipedia \cite{wikidump}, Nalibali \cite{nalibali}, and the Setswana Bible. In addition, we further increase the available Nalibali data by performing PDF text extraction on previously unutilized children's stories and educational materials.

\paragraph{Data Sources.}
Prior works predominantly focus on certain commonly utilized Setswana corpora, limiting downstream model generalization. Marothodi addresses this by directly targeting a rich selection of diverse, manually curated sources. We identify web sources containing Setswana text and develop individual programs to scrape the data as needed. We found many of these sources contain previously underutilized Setswana text in the form of PDFs, making text extraction an important part of Marothodi. We extract text from less readily accessible collections which include educational material \cite{nect2024, thutong, livelingua}, governmental documents~\footnote{\url{https://www.parliament.gov.bw/index.php?}}, and rare books \cite{raretswanabooks}. 

Furthermore, we include miscellaneous individual documents such as the Setswana Universal Declaration of Human Rights \cite{humanrights} and the United States Peace Corps' Intro to Spoken Setswana \cite{introtospokensetswana}, the latter of which required additional image processing, optical character recognition (OCR) with Florence-2 \cite{florence2}, and textual reformatting using Llama 3.1 70B.

We scrape the contents of various websites containing either Setswana or code-switched Setswana-English text, such as iAfrika \cite{iafrika}, Setswana Mo Botswana \cite{setswanamobotswana}, Tlhalefang Communications \cite{tlhalefang}, and the University of South Africa \cite{unisa}. We also include a small corpus of parallel text consisting of monolingual English mathematical text translated into code-mixed English and Setswana \cite{setswanamathcodeswitch}.

Moreover, we include five separate corpora from SADiLaR in Marothodi. This includes a corpus of multilingual code-switched soap opera speech \cite{Corpus_multilingual_codeswitched_soap_opera_speech}, transcripts from a high-quality corpus of Setswana text-to-speech data \cite{High_quality_TTS_South_African_languages}, transcripts from the Lwazi Setswana ASR and TTS corpora \cite{Lwazi_Setswana_ASR_corpus, Lwazi_Setswana_TTS_corpus, Lwazi_II_Setswana_TTS_Corpus}, and transcripts from the NCHLT Setswana Auxiliary Speech Corpus \cite{NCHLT_Setswana_Auxiliary_Speech_Corpus}. We also incorporate smaller existing corpora into Marothodi including Bloom-lm \cite{sil_global_ai_bloomlm_2022}, GlotCC \cite{kargaran2024glotcc}, HPLT \cite{aulamo-etal-2023-hplt}, Vuk'uzenzele \cite{lastrucci-etal-2023-preparing, marivate_vukosi_2023_7598540}, OpenSLR SLR32 \cite{van-niekerk-etal-2017}, and MADLAD-400 \cite{madlad400}

Finally, to further increase the size of our corpus, we employ machine translation using Meta's \textit{NLLB-200-3.3B} \cite{nllb} to translate subsets of TinyStories \cite{tinystories} to Setswana. We found through evaluating round-trip translations \cite{moon-etal-2020-revisiting} that TinyStories' simplistic vocabulary made for generally higher-quality translations compared to more educational text such as FineWeb-edu \cite{fineweb}.

\subsection{Medupi}

\begin{table*}[t]
\centering
\begin{adjustbox}{max width=\linewidth}
\begin{tabular}{@{}l p{8cm} l p{6cm}@{}} % Changed second column to p{6cm}
\toprule
\textbf{Data Category} & \textbf{Sources} & \textbf{Tokens} & \textbf{Description} \\
\midrule
\multirow[t]{3}{*}{Parallel Texts} & OPUS, Autshumato, MAFAND-MT, PolyNews, CaLMQA, SIB-200, MSFT Terms, xP3x, SADiLaR & 178,017,061 & Setswana-English parallel texts for translation; OPUS heavily filtered  \\
\multirow[t]{3}{*}{Translated Corpora} & OpenHermes 2.5, WildChat 1M, Dolly, Magpie Ultra, UltraChat 200k, The Tome, MURI-IT & 50,844,875 & Machine Translated, LLM Translated  \\
\multirow[t]{3}{*}{Augmented Tasks} & MasakhaNER, NCHLT, Daily News Dikgang & 15,607436 & NER, POS, News Classification, and Lemmatization \\
Synthetic Data & GPT-4o (\textit{FineWeb-seeded}) & 2,030,883 & Synthetic LLM-Generated Text  \\
\bottomrule
\end{tabular}
\end{adjustbox}
\caption{\centering Key Data Types and Curation Methods in Medupi.}
\label{tab:medupi_data_types_fullwidth_revised_v2}
\end{table*}

\paragraph{Motivation}
To our knowledge, no comprehensive instruction-tuning or chat-style dataset exists written in Setswana. One notable exception is MURI-IT \cite{koksal2024muri}, although it is comparatively small with content largely focusing on translation-only tasks. Given the lack of available data to pull from and the costs associated with creating human-written corpora, Medupi places a strong focus on data augmentation. In doing so, we demonstrate that meaningful performance improvements can be made for certain languages without substantial financial costs.

Medupi consists of data sourced in three ways:
\begin{enumerate}
    \item Augmenting existing corpora such as parallel texts, Named Entity Recognition (NER), and Part-of-Speech tagging (POS) to fit the expected user-assistant format.
    \item Translating existing corpora using NLLB 200 3.3B, GPT-4o, Gemini 1.5 Pro, and quantized Llama 3.1 405B.
    \item Synthetic text generation with GPT-4o.
\end{enumerate}

Medupi's largest source of Setswana text is the OPUS corpus \cite{opus}, which in total includes over six million parallel English-Setswana sentences across five corpora \cite{ccmatrix, JMLR:v22:20-1307, ccaligned, xlent, tiedemann-2012-parallel, tiedemann-2012-parallel, Tatoeba}. However, we find including this corpus in its entirety in Medupi tends to yield catastrophic forgetting. We attribute this to OPUS's large size relative to the rest of Medupi, where training a model on such an imbalanced dataset hinders its generalization capabilities. We also identify many low-quality English-Setswana parallel sentences, which may further contribute toward catastrophic forgetting.

To mitigate these issues, we translate every English sequence to Setswana using NLLB 3.3B and calculate the CHRF score, comparing the translated Setswana text and the OPUS Setswana sequence. We then filter the entirety of this corpus to only include sequences with the top 33\% of CHRF scores. In doing so, we yield a subset of OPUS whose parallel sequences have greater levels of agreement with existing translation systems and which we find tend to be of greater quality.

Due to data availability, much of Medupi's data sources target translation tasks. In addition to OPUS, we include Autshumato \cite{groenewald2010processing}, MAFAND-MT's training split \cite{mafand-mt}, PolyNews-Parallel \cite{iana2024news}, CaLMQA \cite{calmqa}, SIB-200 \cite{adelani2023sib200}, Microsoft Terms \cite{msterms}, xP3x \cite{muennighoff2022crosslingual}, MURI-IT \cite{koksal2024muri}, and various SADiLaR corpora \cite{COVID-19_Multilingual_Terminology_2021, Van_Dyk_2021, Puttkammer_Hocking_2021}. Each example is formatted to mimic a user-assistant interaction (e.g. "Can you translate the following from English to Setswana..."). To further discourage overfitting and to ensure prompt diversity, we randomize the language and verbiage in the system prompt, whether the source text is provided before or after the user query, the translation direction, and the language provided in the user query for all datasets where applicable. We found through preliminary results these efforts to be effective in mitigating catastrophic forgetting while improving translation performance.

We perform data augmentation on a variety of other data sources. This includes Daily News Dikgang for news classification \cite{puoberta}, MasakhaNER 2.0 for NER \cite{Adelani2022MasakhaNER2A}, and the SADiLaR NCHLT Setswana Annotated Text Corpora for lemmatization and POS tagging \cite{Puttkammer_Schlemmer_Bekker_2021}. We similarly randomize aspects of the system and user prompt where applicable to ensure data diversity.

While data augmentation is important for data diversity in low-resource scenarios, there are still few non-translation sources we could easily adapt for Medupi. As such, we look toward machine translation - a process which has seen prior success in training language models \cite{2403.13638}. We utilize NLLB 200 3.3B to translate 15,200 examples from OpenHermes-2.5 \cite{OpenHermes2.5} and 5,200 examples from WildChat-1M \cite{wildchat}. To encourage multi-turn conversations, we filter WildChat-1M to include examples with at least three turns, and exclude toxic and non-English content. Both datasets are further filtered after translation to remove sequences with signs of low-quality translations such as repetitive text or certain non-Latin characters.

While machine translation models are useful for translating simpler text, we note some limitations with existing models. First, machine translation models do not preserve the original formatting of the text. Second, we find these models are prone to output the input text verbatim when the input text contains code or other technical jargon, indicating formats such as these may be outside the original training distribution. Together, these limitations make it difficult to take advantage of machine translation models when the input text contains code, bulleted lists, mathematical equations, tabular data, and data of highly technical nature.

To obtain additional high-quality Setswana chat-style data while avoiding these problems, we rely on translation using GPT-4o, Gemini 1.5 Pro, and Llama 3.1 405B. We avoid using a single oracle model and instead opt for several teachers to increase diversity with the hope of increased performance \cite{2408.14960}. Specifically, we use GPT-4o to translate a subset of Dolly \cite{dolly}, Magpie \cite{magpie}, and UltraChat 200k \cite{ding2023enhancing, tunstall2023zephyr}, Gemini 1.5 Pro for a subset of OpenHermes 2.5 \cite{OpenHermes2.5}, and we use both Gemini 1.5 Pro and AWQ INT4 quantized Llama 3.1 405B for separate subsets of The Tome \cite{arcee_ai_2024}. We utilize the same filters as previously discussed for NLLB to remove low-quality translations. In addition, we remove instances where the LLM corrupts the format of the translated conversation, such as additional hallucinated system prompts and incorrect user-assistant turn order. This was significantly more common with Llama 3.1 405B, which we attribute to the high degree of quantization.

To further improve the writing quality of the Pula models and experiment with purely synthetic Setswana text, we utilize \textit{gpt-4o-2024-05-13} \cite{gpt4o} to generate 7,860 pieces of text covering diverse topics and styles. To promote data diversity, we seed this process with a random subset of FineWeb and FineWeb-edu \cite{fineweb} and prompt Llama 3 70B to suggest five synopses of related writings while identifying the five most unique words from each seed text. We then cross-reference these words with Google Research's Setswana-English GATITOS \cite{gatitos} to filter for quality, and construct GPT-4o prompts that combine a system instruction, a randomly selected synopsis, and the corresponding word pairs with a requirement that each Setswana word appears in the output. We find this methodology allows GPT-4o to generate high-quality Setswana writings while maintaining a high degree of diversity between texts.

% \vfill
\begin{table*}[h!]
    \centering
    \begin{adjustbox}{max width=\textwidth, scale=0.9}
    \begin{tabular}{|>{\raggedright}p{3.5cm}|>{\raggedright}p{2.5cm}|>{\raggedright}p{2.5cm}|>{\raggedright}p{2.5cm}|>{\raggedright\arraybackslash}p{2.5cm}|}
    \hline
    \textbf{Parameter} & \textbf{Pula 1B} & \textbf{Pula 3B} & \textbf{Pula 8B} & \textbf{Pula 14B} \\
    \hline
    GPUs & 8 & 8 & 8 & 8 \\
    Max Seq Length & 4096 tokens & 4096 tokens & 4096 tokens & 2048 tokens \\
    LoRA Alpha & 32 & 32 & 32 & 32 \\
    LoRA Dropout & 0.2 & 0.2 & 0.2 & 0.2 \\
    LoRA Rank & 64 & 64 & 64 & 16 \\
    Bias & None & None & None & None \\
    Precision & bf16 & bf16 & bf16 & bf16 \\
    Optimizer & AdamW 8bit & AdamW 8bit & AdamW 8bit & AdamW 8bit \\
    Weight Decay & 0.0 & 0.0 & 0.0 & 0.0 \\
    Warmup Ratio & 0.05 & 0.05 & 0.05 & 0.05 \\
    Learning Rate & 2e-05 & 2e-05 & 2e-05 & 2e-05 \\
    Embed Learning Rate & 8e-06 & 8e-06 & 8e-06 & 8e-06 \\
    LR Scheduler & Cosine & Cosine & Cosine & Cosine \\
    Epochs & 3.0 & 3.0 & 3.0 & 3.0 \\
    Packing & \checkmark & \checkmark & \checkmark & \checkmark \\
    Per-Device Batch Size & 1 & 1 & 1 & 1 \\
    Accumulation Steps & 8 & 8 & 8 & 1 \\
    Effective Batch Size & 64 (260k toks) & 64 (260k toks) & 64 (260k toks) & 8 (16k toks) \\
    \hline
    \end{tabular}
    \end{adjustbox}
    \vspace{0.2cm}
    \caption{Training Hyperparameters}
    \label{tab:hyperparams}
\end{table*}

\section{Training}

\paragraph{Training Setup.}
We train all Pula models using DeepSpeed \cite{deepspeed} with ZeRO Stage 3 \cite{zero} and eight NVIDIA H100 GPUs for three epochs on an altered combination of Marothodi and Medupi, totaling approximately 2.3 billion tokens. Our training procedure leverages the Hugging Face \textit{transformers}, \textit{trl}, and \textit{peft} libraries. We develop Pula via continued pre-training, where each model is trained with a large warmup ratio and a lower embedding learning rate to ensure robust learning without catastrophic forgetting.

To train the Pula suite of LLMs effectively while reducing computational demands, we apply Low Rank Adaptation (LoRA) \cite{lora} and QLoRA \cite{qlora} across multiple projection matrices including Query, Key, Value, Output, Gate, Up, and Down. In addition, we maintain full-precision training on the language modeling head and embedding layers to achieve improved performance when adapting our models to Setswana. This approach allows Pula to efficiently learn Setswana without prohibitive computational or memory costs, and it helps to reduce the risk of overfitting and catastrophic forgetting \cite{2405.09673}. We train Pula 1B, 3B, and 8B using LoRA on sequences up to 4096 tokens, whereas Pula 14B is trained using 4-bit QLoRA on sequences up to 2048 tokens in length.

\paragraph{Unified Data Mixture.}
Typical LLM training involves a two-phase pre- and post-training approach, where pre-training or continued pre-training is followed by supervised fine-tuning (SFT) \cite{10.5555/3600270.3602281}. Instead, we train on a mixture of raw text and instruction data by combining Marothodi, our webtext corpus, with Medupi, our instruction-tuning dataset. This dual training strategy not only mitigates the increased computational overhead associated with separate pre-training and post-training phases, but also allows Pula to benefit from the SFT and reinforcement learning benefits already present in the post-trained Llama and Qwen models.

\paragraph{Augmenting Setswana Reasoning and Multilingual Capabilities.}
We found through preliminary results Pula's reasoning performance to be subpar, presumably given the lack of available reasoning data in Medupi. To further increase the representation of Setswana reasoning text in our training data, we selectively duplicate our synthetic LLM datasets (Dolly, Magpie, Ultrachat, OpenHermes, The Tome) three times. To further encourage reasoning, to maintain English capabilities, and to encourage cross-lingual transfer across English and other languages, we incorporate a 15\% mixture of additional datasets, including OpenHermes 2.5 \cite{OpenHermes2.5}, Magpie Pro MT 300k \cite{magpie}, Aya Dataset \cite{aya}, and Inkuba Instruct \cite{inkuba}.

\newcolumntype{M}[1]{>{\centering\arraybackslash}m{#1}}

\begin{table*}[h!]
\centering
\renewcommand{\arraystretch}{1.35}
\begin{adjustbox}{max width=\textwidth}
\begin{tabular}{|M{4.5cm}|M{1.5cm}|*{3}{M{1cm}M{1cm}|M{1cm}M{1cm}|}} 
\hline
\multirow{3}{*}{\textbf{Model}} & \multirow{3}{*}{\textbf{Avg.}} & \multicolumn{4}{c|}{\textbf{MAFAND-MT}} & \multicolumn{4}{c|}{\textbf{Lego-MT}} & \multicolumn{4}{c|}{\textbf{FLORES-200}} \\
\cline{3-14}
 & & \multicolumn{2}{c|}{\textbf{eng-tsn}} & \multicolumn{2}{c|}{\textbf{tsn-eng}} & \multicolumn{2}{c|}{\textbf{eng-tsn}} & \multicolumn{2}{c|}{\textbf{tsn-eng}} & \multicolumn{2}{c|}{\textbf{eng-tsn}} & \multicolumn{2}{c|}{\textbf{tsn-eng}} \\
\cline{3-14}
 & & \textbf{CHRF} & \textbf{BLEU} & \textbf{CHRF} & \textbf{BLEU} & \textbf{CHRF} & \textbf{BLEU} & \textbf{CHRF} & \textbf{BLEU} & \textbf{CHRF} & \textbf{BLEU} & \textbf{CHRF} & \textbf{BLEU} \\
\hline
Llama 3 Instruct (8B) & 11.76 & 22.90 & 2.60 & 29.52 & 5.73 & 11.02 & 1.69 & 11.61 & 1.70 & 20.78 & 2.37 & 27.04 & 4.15 \\
Llama 3 Instruct (70B) & 15.48 & 34.97 & 7.26 & 37.33 & 6.42 & 11.37 & 0.58 & 11.93 & 0.75 & 31.47 & 5.50 & 33.10 & 5.07 \\
Aya 23 (8B) & 7.31 & 14.85 & 0.74 & 17.17 & 1.82 & 8.49 & 0.94 & 9.52 & 1.25 & 14.19 & 0.86 & 16.56 & 1.33 \\
Aya 23 (35B) & 9.77 & 14.95 & 0.88 & 5.96 & 28.88 & 8.37 & 0.89 & 10.16 & 1.19 & 13.05 & 0.63 & 27.09 & 5.13 \\
LLaMAX3 (8B) & 11.59 & 23.39 & 2.00 & 27.55 & 5.62 & 11.57 & 1.29 & 13.21 & 1.98 & 21.51 & 1.64 & 25.12 & 4.14 \\
MADLAD-400 MT (10B) & 17.16 & 22.06 & 6.07 & 34.86 & 14.06 & 16.85 & 9.66 & 22.33 & 14.60 & 19.39 & 3.51 & 31.40 & 11.14 \\
NLLB-200 (3.3B) & 28.46 & \textbf{57.64} & 28.15 & 46.99 & 20.66 & 23.76 & \textbf{13.53} & 17.10 & 4.36 & \textbf{50.15} & \textbf{21.73} & 41.30 & 16.16 \\
\hline
GPT-4o & 30.64 & 51.07 & 23.08 & \textbf{60.91} & \textbf{35.28} & 20.62 & 6.33 & 19.57 & 8.22 & 45.11 & 17.41 & \textbf{52.71} & \textbf{27.40} \\
GPT-4o Mini & 25.33 & 40.12 & 14.20 & 53.11 & 26.18 & 21.04 & 9.00 & 21.16 & 8.22 & 35.29 & 10.05 & 45.98 & 19.59 \\
Gemini 1.5 Pro & 30.67 & 55.71 & 26.29 & 58.87 & 34.10 & 18.10 & 3.99 & 17.20 & 4.72 & 49.81 & 21.59 & 51.16 & 26.49 \\
Gemini 1.5 Flash & 26.50 & 46.98 & 16.20 & 55.01 & 29.66 & 18.23 & 4.42 & 17.42 & 3.65 & 42.41 & 11.87 & 48.51 & 23.60 \\
\hline
Pula-1B & 16.35 & 28.06 & 5.65 & 38.21 & 13.69 & 15.25 & 3.04 & 16.21 & 4.40 & 28.18 & 4.65 & 31.03 & 7.84 \\
Pula-3B & 23.05 & 38.80 & 11.41 & 50.41 & 25.15 & 17.71 & 3.80 & 19.35 & 7.17 & 36.00 & 8.90 & 41.57 & 16.31 \\
Pula-8B & 32.24 & 53.79 & 26.81 & 57.06 & 32.43 & 23.92 & 12.02 & 23.33 & 14.91 & 47.38 & 20.14 & 49.55 & 25.55 \\
Pula-14B & \textbf{33.07} & 55.57 & \textbf{28.48} & 57.31 & 31.94 & \textbf{24.90} & 12.07 & \textbf{24.34} & \textbf{17.40} & 47.78 & 21.00 & 49.98 & 26.11 \\
\hline
\end{tabular}
\end{adjustbox}
\caption{\centering{Translation performance across open and closed models on the MAFAND-MT, Lego-MT, and FLORES-200 benchmarks. We report CHRF and BLEU scores for English-Setswana and Setswana-English translation, as well as average overall score.}}
\label{tab:translation_performance}
\end{table*}

\section{Evaluation}

We evaluate our models on a variety of tasks in both Setswana and English, including translation, natural language understanding, multiple choice question-answering, and mathematical reasoning. All local evaluations are performed using bf16 precision using vLLM \cite{kwon2023efficient}.

\paragraph{Translation.}
To measure translation performance, we evaluate on the MAFAND-MT \cite{mafand-mt}, Lego-MT \cite{lego-mt}, and FLORES-200 \cite{nllb} benchmarks. These benchmarks cover a variety of translation domains, including news headlines, web articles, and miscellaneous web documents. To gauge performance, we utilize the BLEU \cite{bleu} and CHRF \cite{chrf} metrics. Translation performance results are presented in Table \ref{tab:translation_performance}.

To evaluate Pula's knowledge and reasoning capabilities in Setswana we develop \textbf{MMLU-tsn} and \textbf{GSM8K-tsn} - Setswana translations of the entirety of the test splits of the original Massive Multitask Language Understanding (MMLU) \cite{hendryckstest2021, hendrycks2021ethics} and Grade School Math 8K (GSM8K) \cite{cobbe2021gsm8k} benchmarks. We translate using GPT-4o and Gemini 1.5 Pro, respectively. We acknowledge this method's reliance on translation systems to translate this task, especially given its technical nature, is likely to suffer from "translationese" and other errors \cite{mmlulostintranslation}. Many sequences may be incorrectly translated, biased, or otherwise impossible to solve without lucky guessing. However, we find these translated benchmarks to still be a useful proxy for a model's performance in reasoning, knowledge, and instruction-following capabilities. These results are provided in Table \ref{tab:performance}.

We find the Pula series excels at translating between English and Setswana, with both Pula 14B and Pula 8B on average outperforming significantly larger frontier models such as GPT-4o and Gemini 1.5 Pro. These models exceed the performance of all tested open LLMs as well as the NLLB 200 and MADLAD 400 machine translation models. We note the Pula series tends to yield higher quality translations when translating from Setswana to English - a trend we see in other tested LLMs but not in machine translation models.

\paragraph{Reasoning.}
We evaluate reading comprehension using Meta's Belebele benchmark, a corpus of multiple-choice questions regarding passages sourced from FLORES-200. We evaluate world knowledge and question-answering via the MMLU and MMLU-tsn benchmarks, containing multiple-choice questions on topics such as mathematics, computer science, law, and more. Last, we evaluate mathematical reasoning via GSM8K and GSM8K-tsn, which consist of open-ended grade-school math word problems.

\begin{table*}[h!]
\centering
\renewcommand{\arraystretch}{1.3}
\begin{adjustbox}{max width=\textwidth, scale=0.946}
\begin{tabular}{|M{4.5cm}|M{1.5cm}|*{6}{M{0.85cm}|}}
\hline
\multirow{2}{*}{\textbf{Model}} & \multirow{2}{*}{\textbf{Avg.}} & \multicolumn{2}{c|}{\textbf{Belebele}} & \multicolumn{2}{c|}{\textbf{MMLU}} & \multicolumn{2}{c|}{\textbf{GSM8K}} \\
\cline{3-8}
 &  & \textbf{tsn} & \textbf{eng} & \textbf{tsn} & \textbf{eng} & \textbf{tsn} & \textbf{eng} \\
\hline
Llama 3.1 Instruct 8B & 50.32 & 31.04 & 89.67 & 27.73 & 63.75 & 6.61 & 83.09 \\
Llama 3.1 Instruct 70B & 67.69 & 50.69 & 96.00 & 39.95 & 84.28 & 40.24 & 95.00 \\
Aya 23 8B & 32.23 & 30.33 & 62.00 & 26.09 & 44.58 & 2.05 & 28.35 \\
Aya 23 35B & 39.99 & 29.16 & 80.85 & 29.29 & 58.42 & 2.81 & 39.42 \\
LLaMAX3 8B & 27.67 & 30.53 & 68.53 & 25.53 & 41.13 & 0.23 & 0.07 \\
\hline
GPT-4o & 80.59 & 75.33 & \textbf{96.67} & 56.32 & \textbf{87.48} & 72.59 & 95.14 \\
GPT-4o Mini & 66.42 & 46.78 & 94.78 & 40.97 & 80.85 & 41.84 & 93.32 \\
Gemini 1.5 Pro & \textbf{81.26} & \textbf{75.78} & 96.11 & \textbf{60.33} & 86.35 & \textbf{73.80} & \textbf{95.21} \\
Gemini 1.5 Flash & 74.52 & 64.67 & 94.11 & 49.37 & 80.21 & 65.22 & 93.55 \\
\hline
Pula 1B & 21.06 & 27.03 & 36.58 & 24.38 & 28.58 & 3.49 & 6.29 \\
Pula 3B & 35.35 & 30.93 & 50.73 & 30.76 & 44.03 & 11.39 & 44.28 \\
Pula 8B & 59.81 & 51.24 & 85.43 & 38.66 & 61.86 & 43.28 & 78.39 \\
Pula 14B & 69.40 & 66.02 & 91.50 & 46.15 & 74.27 & 57.36 & 81.12 \\
\hline
\end{tabular}
\end{adjustbox}
\vspace{0.2cm}
\caption{\centering{Performance comparison of different large language models on the Belebele, MMLU, and GSM8K benchmarks. For Setswana, we evaluate using the Setswana Belebele subset, MMLU-tsn, and GSM8K-tsn. For English, we evaluate using the English Belebele split, MMLU, and GSM8K.}}
\label{tab:performance}
\end{table*}

We find Pula 14B to, on average, outperform GPT-4o Mini as well as all tested open models across Setswana and English reasoning tasks. Pula 8B outperforms or is competitive with Llama 3.1 70B on all three benchmarks, and Pula 14B significantly outperforms Llama 3.1 70B on all Setswana benchmarks. In addition, we find Pula's performance steadily grows with scale in both Setswana and English tasks; a trend displayed in the performance of existing Llama and Aya models.

However, we do note the roughly equivalent performance of Aya 23 8B and 35B on Setswana Belebele. Given the lack of intentional Setswana data present in Aya 23's training corpus, this indicates reading comprehension may be a less transferable skill for LLMs when working with largely unseen languages compared to question-answering and mathematical reasoning. We also note large discrepancies between Setswana and English versions of benchmarks, such as MMLU and MMLU-tsn. We attribute these differences to accumulative errors during translation such as ambiguous or incorrect wording, impossible questions, or damaging modifications to the correct answer.

\newpage
\section{Conclusion}

In this work we introduce Pula, the first series of large language models tailored for Setswana. Our models demonstrate significantly improved translation and reasoning performance, rivaling models much larger than themselves on Setswana reading comprehension, question-answering, and mathematical reasoning tasks while retaining existing performance on English tasks. Pula exceeds in translating between Setswana and English, with Pula 8B and 14B on average outperforming GPT-4o and Gemini 1.5 Pro, with Pula 14B also outperforming GPT-4o-Mini in Setswana reasoning tasks. We introduce Marothodi, the largest-ever single corpus of raw Setswana text, and Medupi, the first-ever comprehensive Setswana instruction-tuning dataset. We develop and release MMLU-tsn and GSM8K-tsn, Setswana translations of the MMLU and GSM8K benchmarks translated using GPT-4o and Gemini 1.5 Pro. Our results indicate there may be significant performance gains not yet reached in other languages which may be available using existing underutilized data and synthetic data generation. To support future NLP research and production use cases, we release model weights, data, data curation code, benchmarks, training and evaluation code, and training logs.

\newpage

\section{Limitations}

A foundational source for Medupi is translated Setswana instructions and synthetic data using NLLB 200 3.3B, GPT-4o, Gemini 1.5 Pro, and Llama 3.1 405B. The quality of these translations directly influences the quality of much of Medupi and Pula's downstream performance on Setswana tasks. Any inaccuracies, biases, or nuances lost during translation may propagate into the training data and even become more pronounced \cite{2309.00770}.

On a similar note, utilizing translations for benchmarking may introduce "translationese", such as direct translations rather than natural language \cite{2403.13638}. These errors may distort the benchmark's authenticity and reduce the number of answerable questions with corresponding correct answers. These errors may be especially the case for MMLU-tsn and GSM8K-ts, where certain domain-specific vocabulary may not have direct Setswana equivalents.

While Pula demonstrates improved performance compared to existing open source models of its size, there is still significant room for improvement. Further experiments involving additional data translation and filtering at scale, curating human-made chat data, incorporating additional languages, and incorporating additional training methodologies such as multi-stage training, annealing, model merging, and reinforcement learning may allow for increased performance. We hope Pula lays the groundwork for this future work and actively encourage research in these directions.

Last, despite extensive efforts to curate comprehensive corpora of Setswana text, certain cultural and contextual elements may be underrepresented. Dialectical variations, cultural narratives, or region-specific terms, phrases, or other terminology may be comparatively sparse in these corpora. This may lead to models that are less effective in certain situations that require additional cultural insight or sensitivity \cite{mousi2024aradicebenchmarksdialectalcultural}. Addressing these cultural nuances remains an ongoing challenge and an area for future research to ensure language models are properly culturally knowledgeable and configurable for Setswana speakers.

\newpage
\section{Acknowledgements}

We would like to extend our gratitude to the OpenAI team for their invaluable support and for granting us the opportunity to utilize their models. Our appreciation also goes to Trelis Research for their generous financial backing. Additionally, we are deeply thankful to Dr. Jacob Sorber and Professor Carrie Russell of Clemson University, and Dr. Srinath Doss of Botho University. This work would not be possible without your guidance and support. This work is the result of a collaboration that was facilitated by the National Science Foundation under Award CNS 1453607. Any opinions, findings, and conclusions or recommendations expressed in this material are those of the authors and do not necessarily reflect the views of the National Science Foundation.

\section{References}
\bibliography{references}

\end{document}